\documentclass[11pt,a4paper]{article}
\usepackage[hyperref]{naaclhlt2019}
\usepackage{times}
\usepackage{latexsym}

\usepackage{url}
\usepackage{graphicx}
\usepackage{amsmath}
\usepackage{algorithm}
\usepackage{algpseudocode}
\usepackage{mathrsfs}
\usepackage{multirow}
\aclfinalcopy % Uncomment this line for the final submission
\def\aclpaperid{1396} %  Enter the acl Paper ID here

\title{Learning to Decipher Hate Symbols}

\author{Jing Qian, Mai ElSherief, Elizabeth Belding, William Yang Wang\\
  Department of Computer Science \\
  University of California, Santa Barbara\\
  Santa Barbara, CA 93106 USA\\
  {\tt \{jing\_qian,mayelsherif,ebelding,william\}@cs.ucsb.edu} \\}

\date{}

\begin{document}
\maketitle
\begin{abstract}
Existing computational models to understand hate speech typically frame the problem as a simple classification task, bypassing the understanding of hate symbols (e.g., \textit{14 words}, \textit{kigy}) and their secret connotations. In this paper, we propose a novel task of deciphering hate symbols. To do this, we leverage the Urban Dictionary and collected a new, symbol-rich Twitter corpus of hate speech. We investigate neural network latent context models for deciphering hate symbols. More specifically, we study Sequence-to-Sequence models and show how they are able to crack the ciphers based on context. Furthermore, we propose a novel Variational Decipher and show how it can generalize better to unseen hate symbols in a more challenging testing setting.  
\end{abstract}

\section{Introduction}

The statistics are sobering. The Federal Bureau of Investigation of United States\footnote{https://www.fbi.gov/news/stories/2016-hate-crime-statistics} reported over 6,000 criminal incidents motivated by bias against race, ethnicity, ancestry, religion, sexual orientation, disability, gender, and gender identity during 2016. The most recent 2016 report shows an alarming 4.6\% increase, compared with 2015 data\footnote{https://www.fbi.gov/news/stories/2015-hate-crime-statistics-released}. In addition to these reported cases, thousands of Internet users, including celebrities, are forced out of social media due to abuse, hate speech, cyberbullying, and online threats. 
%\textcolor{blue}{ [Add reference here]}
While such social media data is abundantly available, the broad question we are asking is---What can machine learning and natural language processing do to help and prevent online hate speech?

The vast quantity of hate speech on social media can be analyzed to study online abuse. In recent years, there has been a growing trend of developing computational models of hate speech. However, the majority of the prior studies focus solely on modeling hate speech as a binary or multiclass classification task \cite{djuric2015hate,waseem2016hateful,burnap2016us,wulczyn2017ex,pavlopoulos2017deeper}.
 
\begin{figure}[t]
\centering
\includegraphics[width=0.47\textwidth]{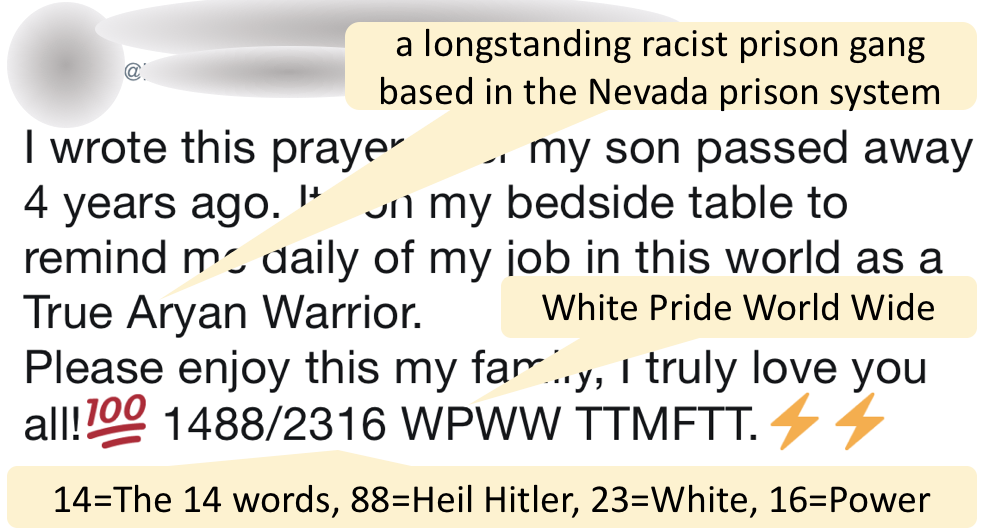}
\caption{An example tweet with hate symbols.}
\label{fig:tweet}
\end{figure}

While developing new features for hate speech detection certainly has merits, we believe that understanding hate speech requires us to design computational models that can decipher hate symbols that are commonly used by hate groups. Figure~\ref{fig:tweet} shows an example usage of hate symbols in an otherwise seemingly harmless tweet that promotes hate. For example, \textit{Aryan Warrior} is a longstanding racist prison gang based in the Nevada prison system. \textit{WPWW} is the acronym for \textit{White Pride World Wide}. The hate symbols \textit{1488} and \textit{2316} are more implicit. \textit{14} symbolizes the 14 words: \textit{``WE MUST SECURE THE EXISTENCE OF OUR PEOPLE AND A FUTURE FOR WHITE CHILDREN''}, spoken by members of the Order neo-Nazi movement. \textit{H} is the 8th letter of the alphabet, so \textit{88}=\textit{HH}=\textit{Heil Hitler}. Similarly, \textit{W} is the 23rd and \textit{P} is the 16th letter of the alphabet, so \textit{2316}=\textit{WP}=\textit{White Power}. \looseness+1
%~\william{Jing, can you find a hate speech tweet example that looks seemingly harmless, but uses hate symbols that truly promote hate? Also, look at Ke Ni's paper and show a plain text to let people know what you task is.} 

In this work, we propose the first models for deciphering hate symbols. We investigate two families of neural network approaches: the Sequence-to-Sequence models~\cite{sutskever2014sequence,cho2014learning} and a novel Variational Decipher based on the Conditional Variational Autoencoders~\cite{kingma2013auto,sohn2015learning,larsen2016autoencoding}. We show how these neural network models are able to guess the meaning of hate symbols based on context embeddings and even generalize to unseen hate symbols during testing. Our contributions are three-fold:
\begin{itemize}
\vspace*{-0.05in}
\item We propose a novel task of learning to decipher hate symbols, which moves beyond the standard formulation of hate speech classification settings.
\vspace*{-0.05in}
\item We introduce a new, symbol-rich tweet dataset for developing computational models of hate speech analysis, leveraging the Urban Dictionary.
\vspace*{-0.05in}
\item We investigate a sequence-to-sequence neural network model and show how it is able to encode context and crack the hate symbols. We also introduce a novel Variational Decipher, which generalizes better in a more challenging setting. 
\end{itemize}
In the next section, we outline related work in text normalization, machine translation, conditional variational autoencoders, and hate speech analysis. 
In Section~\ref{sec:data}, we introduce our new dataset for deciphering hate speech. Next, in Section~\ref{sec:method}, we describe the design of two neural network models for the decipherment problem. Quantitative and qualitative experimental results are presented in Section~\ref{sec:exp}.
Finally, we conclude in Section~\ref{sec:conclusion}.

\section{Related Work} 
\subsection{Text Normalization in Social Media }
The proposed task is related to text normalization focusing on the problems presented by user-generated content in online sources, such as misspelling, rapidly changing out-of-vocabulary slang, short-forms and acronyms, punctuation errors or omissions, etc.  These problems usually appear as out-of-vocabulary words. Extensive research has focused on this task \cite{beaufort2010hybrid,liu2011insertion,gouws2011unsupervised,han2011lexical,han2012automatically,liu2012broad,chrupala2014normalizing}. However, our task is different from the general text normalization in social media in that instead of the out-of-vocabulary words, we focus on the symbols conveying hateful meaning. These hate symbols can go beyond lexical variants of the vocabulary and thus are more challenging to understand.
\subsection{Machine Translation}
An extensive body of work has been dedicated to machine translation.~\citet{knight2006unsupervised} study a number of natural language decipherment problems using unsupervised learning.~\citet{ravi2011deciphering} further frame the task of machine translation as decipherment and tackle it without parallel training data. Machine translation using deep learning (Neural Machine Translation) has been proposed in recent years.~\citet{sutskever2014sequence} and~\citet{cho2014learning} use Sequence to Sequence (Seq2Seq) learning with Recurrent Neural Networks (RNN).~\citet{bahdanau2014neural} further improve translation performance using the attention mechanism. Google's Neural Machine Translation System (GNMT) employs a deep attentional LSTM network with residual connections~\cite{wu2016google}. Recently, machine translation techniques have been also applied to explain non-standard English expressions~\cite{ni2017learning}. However, our deciphering task is not the same as machine translation in that hate symbols are short and cannot be modeled as language. 

Our task is more closely related to ~\cite{hill2016learning} and ~\cite{noraset2017definition}. ~\citet{hill2016learning} propose using neural language embedding models to map the dictionary definitions to the word representations, which is the inverse of our task. ~\citet{noraset2017definition} propose the definition modeling task. However, in their task, for each word to be defined, its pre-trained word embedding is required as an input, which is actually the prior knowledge of the words. However, such kind of prior knowledge is not available in our decipherment task. Therefore, our task is more challenging and is not simply a definition modeling task.

\subsection{Conditional Variational Autoencoder }
Unlike the original Seq2Seq model that directly encodes the input into a latent space, the Variational Autoencoder (VAE)~\cite{kingma2013auto} approximates the underlying probability distribution of data. VAE has shown promise in multiple generation tasks, such as handwritten digits~\cite{kingma2013auto,salimans2015markov}, faces~\cite{kingma2013auto,rezende2014stochastic}, and machine translation~\cite{zhang2016variational}. Conditional Variational Autoencoder~\cite{larsen2016autoencoding,sohn2015learning} extends the original VAE framework by incorporating conditions during generation. In addition to image generation, CVAE has been successfully applied to some NLP tasks. For example,~\citet{zhao2017learning} apply CVAE to dialog generation, while~\citet{guu2018generating} use CVAE for sentence generation.

\subsection{Hate Speech Analysis}
Closely related to our work are~\citet{pavlopoulos2017deeper,gao2017recognizing}. \citet{pavlopoulos2017deeper} build an RNN supplemented by an attention mechanism that outperforms the previous
state of the art system in user comment moderation~\cite{wulczyn2017ex}.~\citet{gao2017recognizing} propose a weakly-supervised approach that jointly trains a slur learner and a hate speech classifier. While their work contributes to the automation of harmful content detection and the highlighting of suspicious words, our work builds upon these contributions by providing a learning mechanism that deciphers suspicious hate symbols used by communities of hate to bypass automated content moderation systems. 

\section{Dataset}
\label{sec:data}
In this section, we describe the dataset we collected for hate symbol decipherment.
\subsection{Hate Symbols}
We first collect hate symbols and the corresponding definitions from the Urban Dictionary. Each term with one of the following hashtags: \textit{\#hate, \#racism, \#racist, \#sexism, \#sexist, \#nazi} is selected as a candidate and added to the set $S_0$. We collected a total of 1,590 terms. Next, we expand this set by different surface forms using the Urban Dictionary API.
%~\textcolor{blue}{[Reference needed here]}. 
For each term $s_i$ in set $S_0$, we obtain a set of terms $R_i$ that have the same meaning as $s_i$ but with different surface forms. For example, for the term \textit{brown shirt}, there are four terms with different surface forms: \textit{brown shirt, brown shirts, Brownshirts, brownshirt}. Each term in $R_i$ has its own definition in Urban Dictionary, but since these terms have exactly the same meaning, we select a definition $d_i$ with maximum upvote/downvote ratio for all the terms in $R_i$. For example, for each term in the set $R_i$=\textit{\{brown shirt, brown shirts, Brownshirts, brownshirt\}}, the corresponding definition is \textit{``Soldiers in Hitler's storm trooper army, SA during the Nazi regime...''}
%~\textcolor{blue}{[Not clear what is SA'']}}. 
After expanding, we obtain 2,105 distinct hate symbol terms and their corresponding definitions. On average, each symbol consists of 9.9 characters, 1.5 words. Each definition consists of 96.8 characters, 17.0 words.

\subsection{Tweet Collection}
For each of the hate symbols, we collect all tweets from 2011-01-01 to 2017-12-31 that contain exactly the same surface form of hate symbol in the text. Since we only focus on hate speech, we train an SVM ~\cite{cortes1995support} classifier to filter the collected tweets. The SVM model is trained on the dataset published by ~\citet{waseem2016hateful}. Their original dataset contains three labels: Sexism, Racism, and None. Since the SVM model is used to filter the non-hate speech, we merge the instances labeled as sexism and racism, then train the SVM model to do binary classification. After filtering out all the tweets classified as non-hate, our final dataset consists of 18,667 (tweet, hate symbol, definition) tuples.

\section{Our Approach} 
\label{sec:method}
We formulate hate symbol deciphering as the following equation:
\begin{align}
Obj=\sum_{(u,s,d^*)\in X}\log p(d^*|(u,s))
\end{align}
$X$ is the dataset, $(u,s,d^*)$ is the (tweet, symbol, definition) tuple in the dataset. The inputs are the tweet and the hate symbol in this tweet. The output is the definition of the symbol. Our objective is to maximize the probability of the definition given the (tweet, symbol) pair. This objective function is very similar to that of machine translation. So we first try to tackle it based on the Sequence-to-Sequence model, which is commonly used in machine translation.
\subsection{Sequence-to-Sequence Model}\label{sec:seq2seq}

\begin{figure}[t]
\centering
\includegraphics[width=0.8\linewidth]{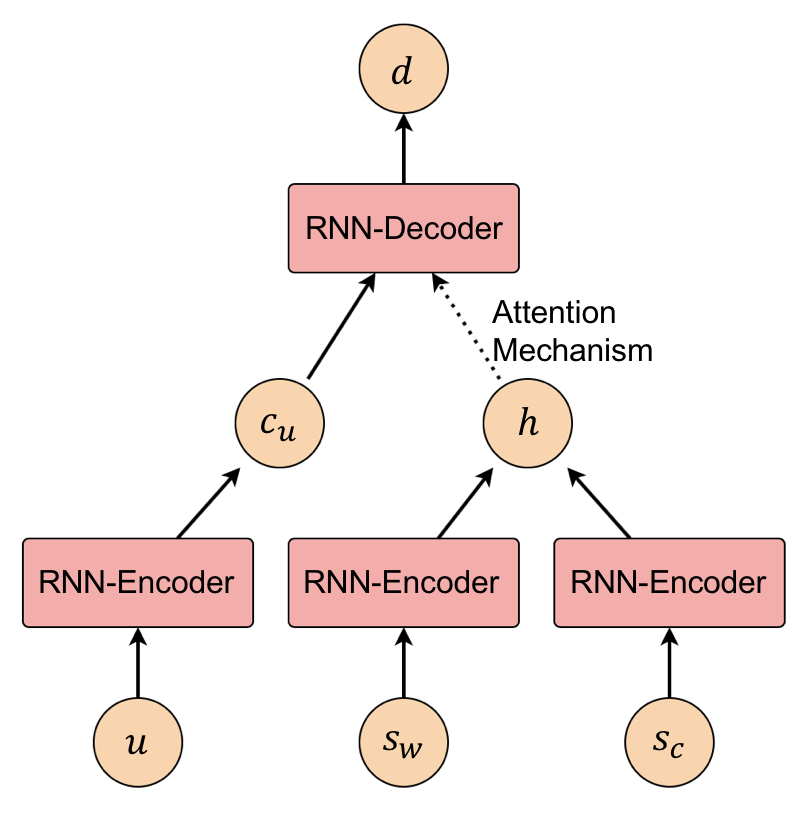}
\caption{Our Seq2Seq model. $u$, $s_w$ are the word embeddings of the tweet text and hate symbol. $s_c$ is the character embedding of the symbol. $c_u$ is the encoded tweet and $h$ is the concatenated hidden states. $d$ is the generated text. Detailed explanation is in section~\ref{sec:seq2seq}.}
\label{fig:seq2seq}
\end{figure}

We implement an RNN Encoder-Decoder model with attention mechanism based on \citet{bahdanau2014neural}. We use GRU ~\cite{cho2014learning} for decoding. However, instead of also using GRU for encoding, we found that LSTM ~\cite{hochreiter1997long} performs better on our task. Therefore, our Seq2Seq model uses LSTM encoders and GRU decoders. An overview of our Seq2Seq model is shown in Figure~\ref{fig:seq2seq}. The computation process is shown as the following equations: 
\begin{align}
c_u,h_u=f_u(u)\\
c_{sw},h_{sw}=f_{sw}(s_w)\\
c_{sc},h_{sc}=f_{sc}(s_c)
\end{align}
$u$ is the word embedding of the tweet text, $s_w$ is the word embedding of the hate symbol, $s_c$ is the character embedding of the symbol. $f_u$, $f_{sw}$, and $f_{sc}$ are LSTM functions. $c_u$, $c_{sw}$, $c_{sc}$ are the outputs of the LSTMs at the last time step and $h_u$, $h_{sw}$, $h_{sc}$ are the hidden states of the LSTMs at all time steps. We use two RNN encoders to encode the symbol, one encodes at the word level and the other one encodes at the character level. The character-level encoded hate symbol is used to provide the feature of the surface form of the hate symbol while the word-level encoded hate symbol is used to provide the semantic information of the hate symbol.  
The hidden states of the two RNN encoders for hate symbols are concatenated:
\begin{align}
h=h_{sw}\oplus h_{sc}
\end{align}
$c_u$ is the vector of encoded tweet text. The tweet text is the context of the hate symbol, which provides additional information during decoding. Therefore, the encoded tweet text it is also fed into the RNN decoder. The detailed attention mechanism and decoding process at time step $t$ are as follows: 
\begin{align}
w_t=\sigma(l_{w}(d_{t-1}\oplus e_{t-1}))\\
a_t=\sum_{i=1}^T{w_{ti}h_i}\\
b_t=\sigma(l_c(d_{t-1}\oplus a_t))\\
o_t,e_t=k(c_u\oplus b_t,e_{t-1})\\
p(d_t|u,s)=\sigma (l_o(o_t))
\end{align}
$w_t$ is the attention weights at time step $t$ and $w_{ti}$ is the $i_{th}$ weight of $w_t$. $d_{t-1}$ is the generated word at last time step and $e_{t-1}$ is the hidden state of the decoder at last time step. $h_i$ is the $i_{th}$ time step segment of $h$. $l_w$, $l_c$, and $l_o$ are linear functions. $\sigma$ is a nonlinear activation function. $k$ is the GRU function. $o_t$ is the output and $e_t$ is the hidden state of the GRU. $p(d_t|u,s)$ is the probability distribution of the vocabulary at time step ${t}$. The attention weights ${w_t}$ are computed based on the decoder's hidden state and the generated word at time step ${t-1}$. Then the computed weights are applied to the concatenated hidden states $h$ of encoders. The result $a_t$ is the context vector for the decoder at time step $t$. The context vector and the last generated word are combined by a linear function $l_c$ followed by a nonlinear activation function. The result $b_t$ is concatenated with the encoded tweet context $c_u$, and then fed into GRU together with the decoder's last hidden state $e_{t-1}$. Finally, the probability of each vocabulary word is computed from $o_t$.

\subsection{Variational Decipher}\label{sec:cvae}

\begin{figure}[t]
\centering
\includegraphics[width=0.8\linewidth]{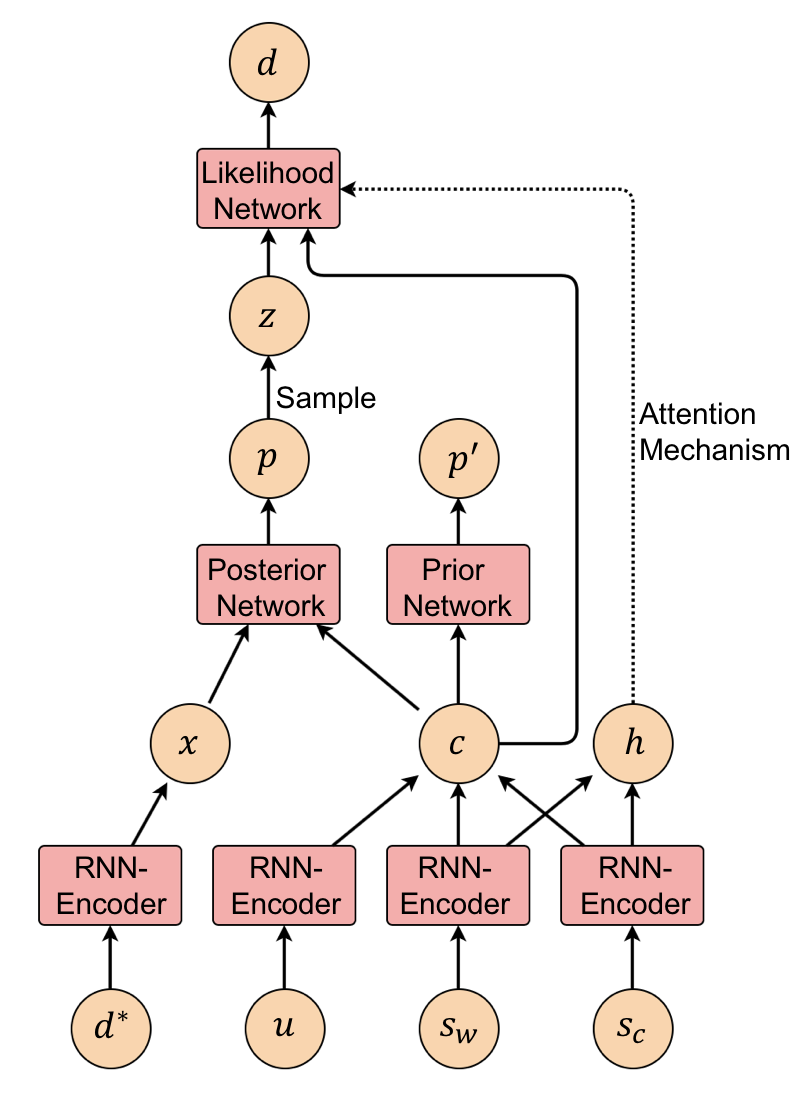}
\caption{The Variational Decipher. Note that this structure is used during training. During testing, the structure is slightly different. $d^*$ is the word embeddings of the definition. $x$ is the encoded definition. $c$ is the concatenation of the encoded tweet and hate symbol. $p$ and $p^\prime$ are output distributions. $z$ is the latent variable. The definitions of other variables are the same as those in Figure~\ref{fig:seq2seq}. Detailed explanation is in section~\ref{sec:cvae}.}
\label{fig:cvae}
\end{figure}
The Variational Decipher is based on the CVAE model, which is another model that can be used to parametrize the conditional probability $p(d^*|(u,s))$ in the objective function (Equation 1). Unlike the Seq2Seq model, which directly parametrizes $p(d^*|(u,s))$, our variational decipher formulates the task as follows: 
\begin{equation}
\begin{split}
Obj&=\sum_{(u,s,d^*)\in X}\log p(d^*|(u,s))\\
&\!=\!\sum_{(u,s,d^*)\in X}\!\!\log\!\int_z p(d^*|z)p(z|(u,s))dz
\end{split}
\end{equation}
where $z$ is the latent variable. $p(d^*|(u,s)$ is written as the marginalization of the product of two terms over the latent space. Since the integration over $z$ is intractable, we instead try to maximize the evidence lower bound (ELBO). Our variational lower bound objective is in the following form:
\begin{equation}
\begin{aligned}
Obj=&E[\log p_\varphi(d^*|z,u,s)]-\\
&D_{KL}[p_\alpha(z|d^*,u,s)||p_\beta(z|u,s)]
\end{aligned}
\end{equation}
where $p_\varphi(d^*|z,u,s)$ is the likelihood, $p_\alpha(z|d^*,u,s)$ is the posterior, $p_\beta(z|u,s)$ is the prior, and $D_{KL}$ is the Kullback-Leibler (KL) divergence.  We use three neural networks to model these three probability distributions.
An overview of our variational decipher is shown in Figure~\ref{fig:cvae}.
We first use four recurrent neural networks to encode the (tweet, symbol, definition) pair in the dataset. Similar to what we do in the Seq2Seq model, there are two encoders for the hate symbol. One is at the word level and the other is at the character level. The encoding of symbols and tweets are exactly the same as in our Seq2Seq model (see Equations 2-4). The difference is that we also need to encode definitions for the Variational Decipher. 
\begin{align}
x,h_d=f_d(d^*)
\end{align}
Here, $f_d$ is the LSTM function. $x$ is the output of the LSTM at the last time step and $h_d$ is the hidden state of the LSTM at all time steps.
The condition vector $c$ is the concatenation of the encoded symbol words, symbol characters, and the tweet text:
\begin{align}
c=c_u\oplus c_{sw}\oplus c_{sc}
\end{align}
We use multi-layer perceptron (MLP) to model the posterior and the prior in the objective function. The posterior network and the prior network have the same structure and both output a probability distribution of latent variable $z$. The only difference is that the input of the posterior network is the concatenation of the encoded definition $x$ and the condition vector $c$ while the input of the prior network is only the condition vector $c$. Therefore, the output of the posterior network $p=p_\alpha(z|d^*,u,s)$ and the output of the prior network $p^\prime=p_\beta(z|u,s)$. By assuming the latent variable $z$ has a multivariate Gaussian distribution, the actual outputs of the posterior and prior networks are the mean and variance: ($\mu$, $\Sigma$) for the posterior network and ($\mu^\prime$, $\Sigma^\prime$) for the prior network. 
\begin{align}
\mu,\Sigma=g(x\oplus c)\\
\mu^\prime,\Sigma^\prime=g^\prime(c)
\end{align}
$g$ is the MLP function of the posterior network and $g^\prime$ is that of the prior network. 
During training, the latent variable $z$ is randomly sampled from the Gaussian distribution $\mathcal N(\mu,\Sigma)$ and fed into the likelihood network. During testing, the posterior network is replaced by the prior network, so $z$ is sampled from $\mathcal N(\mu^\prime,\Sigma^\prime)$. The likelihood network is modeled by an RNN decoder with attention mechanism, very similar to the decoder of our Seq2Seq model. The only difference lies in the input for the GRU. The decoder in our Variational Decipher model is to model the likelihood $p_\varphi(d^*|z,u,s)$, which is conditioned on the latent variable, tweet context, and the symbol. Therefore, for the Variational Decipher, the condition vector $c$ and the sampled latent variable $z$ are fed into the decoder.
\begin{align}
o_t,e_t=k(z\oplus c\oplus b_t,e_{t-1})
\end{align}
$e_{t-1}$ is the hidden state of the RNN decoder at the last time step. $k$ is the GRU function. $o_t$ is its output and $e_t$ is its hidden state. Detailed decoding process and explanations are in section~\ref{sec:seq2seq}.

According to the objective function in Equation 12, the loss function of the Variational Decipher is as follows:
\begin{equation}
\begin{aligned}
\mathcal{L}=&L_{REC}+L_{KL}\\=&E_{z\sim p_\alpha(z|d^*,u,s)}[-\log p_\varphi(d^*|z,u,s)]+\\
&D_{KL}[p_\alpha(z|d^*,u,s)||p_\beta(z|u,s)]
\end{aligned}
\end{equation}
It consists of two parts. The first part $L_{REC}$ is called reconstruction loss. Optimizing $L_{REC}$ can push the sentences generated by the posterior network and the likelihood network closer to the given definitions. The second part $L_{KL}$ is the KL divergence loss. Optimizing this loss can push the output Gaussian Distributions of the prior network closer to that of the posterior network. This means we teach the prior network to learn the same knowledge learned by the posterior network, such that during testing time, when the referential definition $d^*$ is no longer available for generating the latent variable $z$, the prior network can still output a reasonable probability distribution over the latent variable $z$. The complete training and testing process for the Variational Decipher is shown in Algorithm~\ref{alg:algorithm}. $M$ is the predefined maximum length of the generated text. $BCE$ refers to the Binary Cross Entropy loss.

\begin{algorithm}[t]
\small
  \caption{Train \& Test Variational Decipher}\label{alg:algorithm}
  \begin{algorithmic}[1]
  \Function{train}{$U$}
    \State randomly initialize network parameters $\varphi$, $\alpha$, $\beta$;
    \For{$epoch=1, E$}
      \For{$(tweet, symbol, definition)$ in $U$}
        \State get embeddings $u$, $s_w$, $s_c$, $d^*$;
        \State compute $x$, $c$ and $h$ with RNN encoders;
        \State compute $\mu$, $\Sigma$ with the posterior network;
        \State compute $\mu^\prime$, $\Sigma^\prime$ with the prior network;
        \State compute KL-divergence loss $L_{KL}$;
        \State sample $z=reparameterize(\mu,\Sigma)$;
        \State initialize the decoder state $e_0=c$;
        \State $L_{REC}=0$;
        \For{$t=1, M$}
          \State compute attention weights $w_t$;
          \State compute $o_{t}$, $e_{t}$ and $p(d_t|z,u,s)$;
          \State $d_t=indmax(p(d_t|z,u,s))$;
          \State $L_{REC}+=BCE(d_t,d^*_t)$;
          \If {$d_t$==EOS}
          	\State break;
          \EndIf
        \EndFor
        \State update $\varphi$, $\alpha$, $\beta$ on $\mathcal{L}=L_{REC}+L_{KL}$; 
      \EndFor
    \EndFor
   \EndFunction
   \State
   \Function{test}{$V$}
   	\For{$(tweet, symbol, definition)$ in $V$}
    	\State get embeddings $u$, $s_w$, $s_c$;
        \State compute $c$ and $h$ with RNN encoders;
        \State compute $\mu^\prime$, $\Sigma^\prime$ with the prior network;
        \State sample $z=reparameterize(\mu^\prime,\Sigma^\prime)$;
        \State initialize the decoder state $e_0=c$;
        \For{$t=1, M$}
          \State compute attention weights $w$;
          \State compute $o_{t}$, $e_{t}$ and $p(d_t|z,u,s)$;
          \State $d_t=indmax(p(d_t|z,u,s))$;
          \If {$d_t$==EOS}
          	\State break;
          \EndIf
        \EndFor
    \EndFor
   \EndFunction
  \end{algorithmic}
\end{algorithm}

\section{Experiments} 
\label{sec:exp}
\subsection{Experimental Settings}
We use the dataset collected as described in section~\ref{sec:data} for training and testing. We randomly selected 2,440 tuples for testing and use the remaining 16,227 tuples for training. Note that there are no overlapping hate symbols between the training dataset $U$ and the testing dataset $D$. 

We split the 2,440 tuples of the testing dataset $D$ into two separate parts, $D_s$ and $D_d$. $D_s$ consists of 1,681 examples and $D_d$ consists of 759 examples. In the first testing dataset $D_{s}$, although each hate symbol does not appear in the training dataset, the corresponding definition appears in the training dataset. In the second testing dataset $D_d$, neither the hate symbols nor the corresponding definitions appear in the training dataset. We do this split because deciphering hate symbols in these two cases has different levels of difficulty. 

This split criterion means that for each hate symbol in $D_s$, there exists some symbol in the training dataset that has the same meaning but in different surface forms. For example, the hate symbol \textit{wigwog} and \textit{Wig Wog} have the same definition but one is in the training dataset, the other is in the first testing dataset. We assume that such types of hate symbols share similar surface forms or similar tweet contexts. Therefore, the first testing dataset $D_s$ is to evaluate how well the model captures the semantic similarities among the tweet contexts in different examples or the similarities among different surface forms of a hate symbol. 

Deciphering the hate symbols in the second testing dataset $D_d$ is more challenging. Both the unseen hate symbols and definitions require the model to have the ability to accurately capture the semantic information in the tweet context and then make a reasonable prediction. The second testing dataset $D_d$ is used to evaluate how well the model generalizes to completely new hate symbols. 

For the Seq2Seq model, we use negative log-likelihood loss for training. Both models are optimized using Adam optimizer ~\cite{kingma2014adam}.
%~\textcolor{blue}{[Add reference here]}. 
The hyper-parameters of two models are exactly the same. We set the maximum generation length $M=50$. The hidden size of the encoders is 64. The size of the word embedding is 200 and that of character embedding is 100. The word embeddings and character embeddings are randomly initialized. Each model is trained for 50 epochs. We report the deciphering results of two models on three testing datasets $D$, $D_s$ and $D_d$.

\subsection{Experimental Results}
\begin{figure}[t]
\centering
\includegraphics[width=1.0\linewidth]{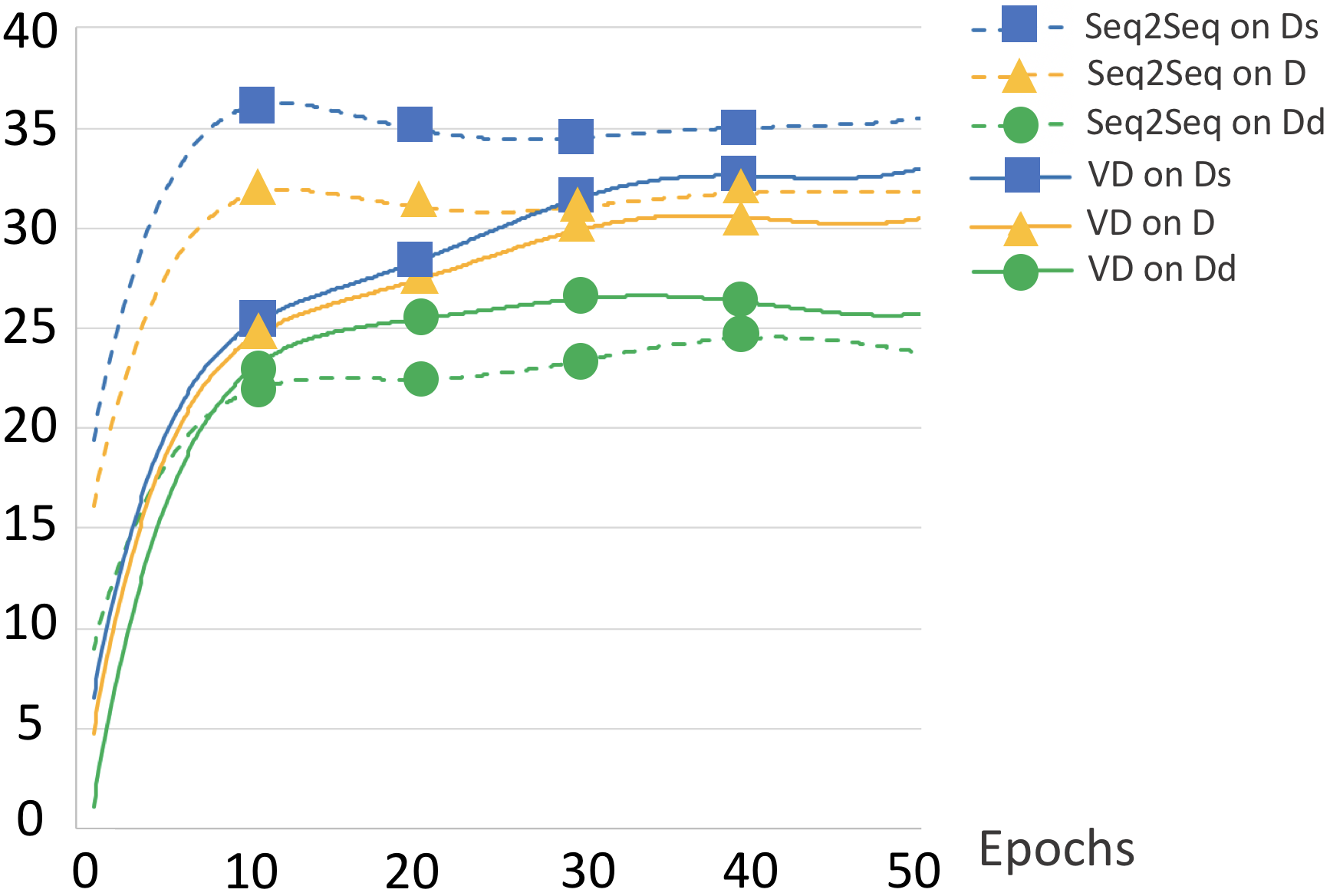}
\caption{BLEU scores of two models on the testing dataset $D$, $D_s$ and $D_d$. The three dotted curves represent the performance of the Seq2Seq model while the three solid curves represent the performance of the Variational Decipher. }
\label{fig:bleu}
\end{figure}

\vspace{3pt}
\noindent{\bf Quantitative Results:}
We use equally weighted BLEU score for up to 4-grams ~\cite{papineni2002bleu}, ROUGE-L ~\cite{lin2004rouge} and METEOR ~\cite{banerjee2005meteor}
%~\textcolor{blue}{[Add year to reference to appear here]}. 
to evaluate the decipherment results. The results are shown in Table~\ref{tab:bleu}. Figure~\ref{fig:bleu} shows the BLEU score achieved by the two models on three testing datasets $D$, $D_s$ and $D_d$ during the training process. Both our Seq2Seq model and Variational Decipher achieve reasonable BLEU scores on the testing datasets. The Seq2Seq model outperforms the Variational Decipher on $D_s$ while Variational Decipher outperforms Seq2Seq on $D_d$. Note that $D_s$ is more than twice the size of $D_d$. Therefore, Seq2Seq outperforms Variational Decipher on the entire testing dataset $D$. The different performance of the two models on $D_s$ and $D_d$ is more obvious in Figure~\ref{fig:bleu}. The gap between the performance of the Seq2Seq model on $D_s$ and $D_d$ is much larger than that between the performance of the Variational Decipher on these two datasets.

\begin{table}[t!]
\centering
\small
\begin{tabular}{|c|c|c|c|c|}
  \hline
    Dataset   &Method& BLEU & ROUGE-L &METEOR  \\
  \hline
  \multirow{2}*{$D_s$} &Seq2Seq &\textbf{37.80} &\textbf{41.05} &\textbf{36.67} \\
  &VD &34.77&32.96&31.03\\
  \hline
  \multirow{2}*{$D_d$} &Seq2Seq &25.44 &12.96 &\textbf{5.54} \\
  &VD &\textbf{28.38} &\textbf{14.01} &5.41\\
  \hline
  \multirow{2}*{$D$} &Seq2Seq &\textbf{33.96} &\textbf{32.32} &\textbf{26.98} \\
  &VD &32.75&27.00&23.16\\
  \hline
\end{tabular}
\caption{The BLEU, ROUGE-L and METEOR scores on testing datasets. VD refers to the Variational Decipher. $D$ is the entire testing dataset. $D_s$ is the first part of $D$ and $D_d$ is the second part. The better results are in bold. }
\label{tab:bleu}
\end{table}

\vspace{3pt}
\noindent{\bf Human Evaluation:}
We employed crowd-sourced workers to evaluate the deciphering results of two models. We randomly sampled 100 items of deciphering results from $D_s$ and another 100 items from $D_d$. Each item composes a choice question and each choice question is assigned to five workers on Amazon Mechanical Turk. In each choice question, the workers are given the hate symbol, the referential definition, the original tweet and two machine-generated plain texts from the Seq2Seq model and Variational Decipher. Workers are asked to select the more reasonable of the two results. In each choice question, the order of the results from the two models is permuted. Ties are permitted for answers. We batch five items in one assignment and insert an artificial item with two identical outputs as a sanity check. The workers who fail to choose ``tie'' for that item are rejected from our test. The human evaluation results are shown in Table \ref{tab:mturk}, which coincide with the results in Table~\ref{tab:bleu} and Figure~\ref{fig:bleu}.

\begin{table}[t!]
\centering
\small
\begin{tabular}{|c|c|c|c|}
  \hline
  Dataset &Seq2Seq Lose & Seq2Seq Win &Tie   \\
  \hline
  {$D_s$} &31.0\% &32.0\% &37.0\% \\
  
  \hline
  {$D_d$} &30.5\% &22.0\% &47.5\% \\
  \hline
\end{tabular}
\caption{The results of human evaluation on two separate testing datasets $D_s$ and $D_d$.  }
\label{tab:mturk}
\end{table}
\begin{figure*}[t]
\centering
\includegraphics[width=0.98\textwidth]{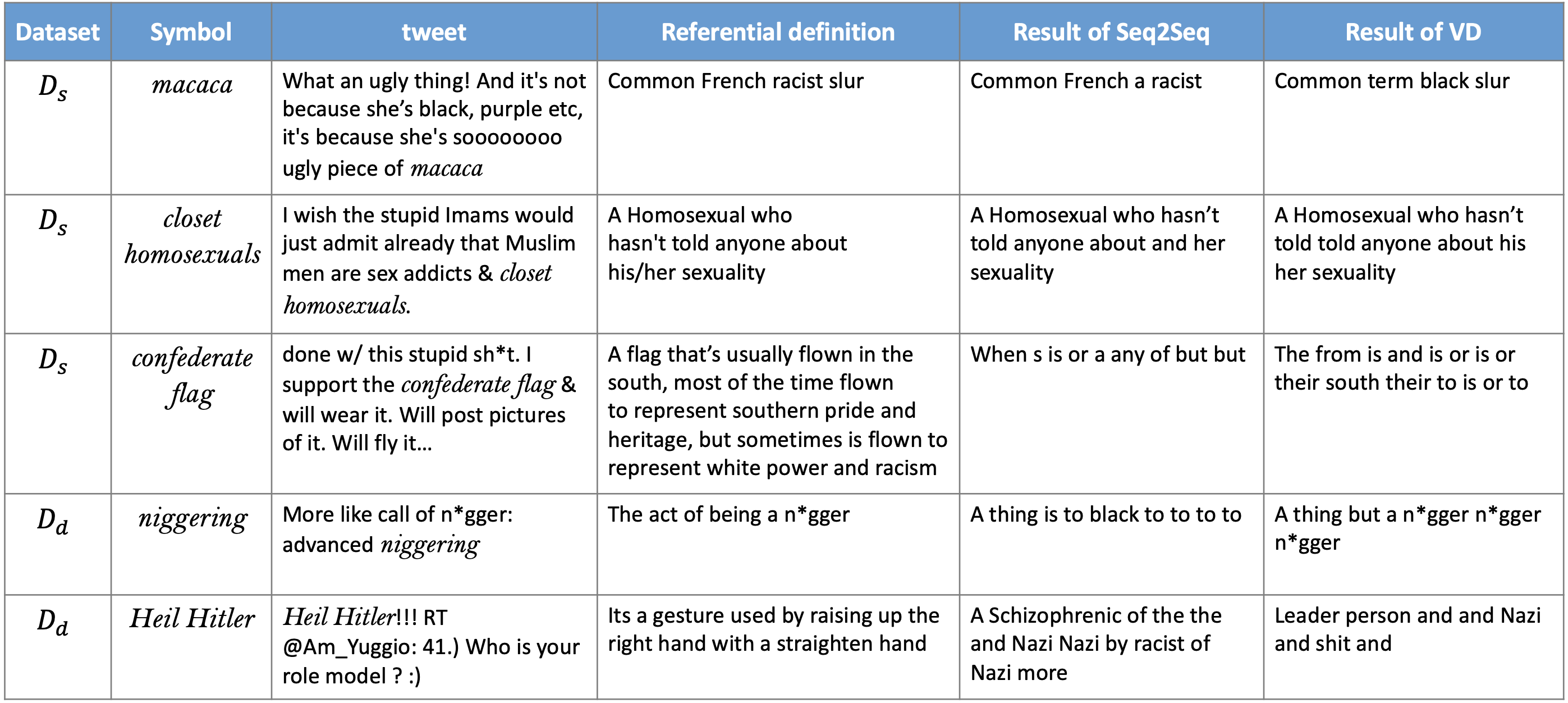}
\caption{Some example errors in the generated results of our Seq2Seq model and Variational Decipher.
}
\label{fig:examples}
\end{figure*}
\vspace{3pt}
\noindent{\bf Discussion:}
When deciphering the hate symbols that have the same definitions as in the training dataset, the model can rely more on the surface forms of hate symbols than the tweet context to make a prediction because usually the hate symbols that share the same definitions also have similar surface forms. However, when it comes to the hate symbols with unseen definitions, simply relying on the surface forms cannot lead to a reasonable deciphering result. Instead, the model should learn the relationships between the context information and the definition of the symbol. Therefore, the different performances of two models on the two testing datasets $D_s$ and $D_d$ indicate that the Seq2Seq model is better at capturing the similarities among different surface forms of a hate symbol, while the Variational Decipher is better at capturing the semantic relationship between the tweet context and the hate symbol. The Sequence-to-Sequence model tries to capture such kinds of relationships by compressing all the context information into a fixed length vector, so its deciphering strategy is actually behavior cloning. On the other hand, the Variational Decipher captures such relationships by explicitly modeling the posterior and likelihood distributions. The modeled distributions provide higher-level semantic information compared to the compressed context, which allows the Variational Decipher to generalize better to the symbols with unseen definitions. This explains why the gap between the performance of the Seq2Seq model on two datasets is larger. \looseness+1

\subsection{Error Analysis}
Figure~\ref{fig:examples}
%~\textcolor{blue}{Is that a figure or a table?} 
shows some example errors of the deciphering results of our Seq2Seq model and Variational Decipher. 
One problem with the deciphering results is that the generated sentences have poor grammatical structure, as shown in Figure~\ref{fig:examples}. This is mainly because the size of our dataset is small, and the models need a much larger corpus to learn the grammar. We anticipate that the generation performance will be improved with a larger dataset.

For the hate symbols in $D_s$, the deciphering results are of high quality when the length of referential definitions are relatively short. An example is \textit{macaca}, a French slur shows in Figure~\ref{fig:examples}. The deciphering result of the Seq2Seq model is close to the referential definition. As to the Variational Decipher, although the result is not literally the same as the definition, the meaning is close. \textit{closet homosexuals} in Figure~\ref{fig:examples} is another example. However, when the length of the referential definition increases, the performance of both models tends to be unsatisfactory, as the third example \textit{confederate flag} shows in Figure~\ref{fig:examples}. Although there exists the symbol \textit{Confederate Flag} with the same definition in the training set, both models fail on this example. One possible reason is that the complexity of generating the referential definition grows substantially with the increasing length, so when the tweet context and the symbol itself cannot provide enough information, the generation model cannot learn the relationship between the symbol and its definition. 

Deciphering hate symbols in $D_d$ is much more challenging. Even for humans, deciphering completely new hate symbols is not a simple task. The two examples in Figure~\ref{fig:examples} show that the models have some ability to capture the semantic similarities. For the symbol \textit{niggering}, the Variational Decipher generates the word \textit{nigger} and Seq2Seq model generates \textit{black}. For \textit{Heil Hitler}, the Variational Decipher generates \textit{leader person} and \textit{Nazi}, while Seq2Seq also generates \textit{Nazi}. Although these generated words are not in the definition, they still make some sense.

\section{Conclusion}
\label{sec:conclusion}
We propose a new task of learning to decipher hate symbols and create a symbol-rich tweet dataset. We split the testing dataset into two parts to analyze the characteristics of the Seq2Seq model and the Variational Decipher. The different performance of these two models indicates that the models can be applied to different scenarios of hate symbol deciphering. The Seq2Seq model outperforms the Variational Decipher for deciphering the hate symbols with similar definitions to that in the training dataset. This means the Seq2Seq model can better explain the hate symbols when Twitter users intentionally misspell or abbreviate common slur terms. On the other hand, the Variational Decipher tends to be better at deciphering hate symbols with unseen definitions, so it can be applied to explain newly created hate symbols on Twitter. Although both models show promising deciphering results, there still exists much room for improvement. 
\bibliography{naaclhlt2019}

\begin{thebibliography}{36}
\expandafter\ifx\csname natexlab\endcsname\relax\def\natexlab#1{#1}\fi

\bibitem[{Bahdanau et~al.(2015)Bahdanau, Cho, and Bengio}]{bahdanau2014neural}
Dzmitry Bahdanau, Kyunghyun Cho, and Yoshua Bengio. 2015.
\newblock Neural machine translation by jointly learning to align and
  translate.
\newblock In \emph{Proceedings of the 3rd International Conference on Learning
  Representations}.

\bibitem[{Banerjee and Lavie(2005)}]{banerjee2005meteor}
Satanjeev Banerjee and Alon Lavie. 2005.
\newblock Meteor: An automatic metric for {MT} evaluation with improved
  correlation with human judgments.
\newblock In \emph{Proceedings of the ACL workshop on intrinsic and extrinsic
  evaluation measures for machine translation and/or summarization}, pages
  65--72.

\bibitem[{Beaufort et~al.(2010)Beaufort, Roekhaut, Cougnon, and
  Fairon}]{beaufort2010hybrid}
Richard Beaufort, Sophie Roekhaut, Louise-Am{\'e}lie Cougnon, and C{\'e}drick
  Fairon. 2010.
\newblock A hybrid rule/model-based finite-state framework for normalizing
  {SMS} messages.
\newblock In \emph{Proceedings of the 48th Annual Meeting of the Association
  for Computational Linguistics}, pages 770--779. Association for Computational
  Linguistics.

\bibitem[{Burnap and Williams(2016)}]{burnap2016us}
Pete Burnap and Matthew~L Williams. 2016.
\newblock Us and them: identifying cyber hate on {Twitter} across multiple
  protected characteristics.
\newblock \emph{EPJ Data Science}, 5(1):11.

\bibitem[{Cho et~al.(2014)Cho, van Merrienboer, Gulcehre, Bahdanau, Bougares,
  Schwenk, and Bengio}]{cho2014learning}
Kyunghyun Cho, Bart van Merrienboer, Caglar Gulcehre, Dzmitry Bahdanau, Fethi
  Bougares, Holger Schwenk, and Yoshua Bengio. 2014.
\newblock Learning phrase representations using {RNN} encoder--decoder for
  statistical machine translation.
\newblock In \emph{Proceedings of the 2014 Conference on Empirical Methods in
  Natural Language Processing (EMNLP)}, pages 1724--1734.

\bibitem[{Chrupa{\l}a(2014)}]{chrupala2014normalizing}
Grzegorz Chrupa{\l}a. 2014.
\newblock Normalizing tweets with edit scripts and recurrent neural embeddings.
\newblock In \emph{Proceedings of the 52nd Annual Meeting of the Association
  for Computational Linguistics (Volume 2: Short Papers)}, volume~2, pages
  680--686.

\bibitem[{Cortes and Vapnik(1995)}]{cortes1995support}
Corinna Cortes and Vladimir Vapnik. 1995.
\newblock Support-vector networks.
\newblock \emph{Machine Learning}, 20(3):273--297.

\bibitem[{Djuric et~al.(2015)Djuric, Zhou, Morris, Grbovic, Radosavljevic, and
  Bhamidipati}]{djuric2015hate}
Nemanja Djuric, Jing Zhou, Robin Morris, Mihajlo Grbovic, Vladan Radosavljevic,
  and Narayan Bhamidipati. 2015.
\newblock Hate speech detection with comment embeddings.
\newblock In \emph{Proceedings of the 24th ACM International Conference on
  World Wide Web}, pages 29--30.

\bibitem[{Gao et~al.(2017)Gao, Kuppersmith, and Huang}]{gao2017recognizing}
Lei Gao, Alexis Kuppersmith, and Ruihong Huang. 2017.
\newblock Recognizing explicit and implicit hate speech using a weakly
  supervised two-path bootstrapping approach.
\newblock In \emph{Proceedings of the Eighth International Joint Conference on
  Natural Language Processing}, pages 774--782.

\bibitem[{Gouws et~al.(2011)Gouws, Hovy, and Metzler}]{gouws2011unsupervised}
Stephan Gouws, Dirk Hovy, and Donald Metzler. 2011.
\newblock Unsupervised mining of lexical variants from noisy text.
\newblock In \emph{Proceedings of the First workshop on Unsupervised Learning
  in NLP}, pages 82--90. Association for Computational Linguistics.

\bibitem[{Guu et~al.(2018)Guu, Hashimoto, Oren, and Liang}]{guu2018generating}
Kelvin Guu, Tatsunori~B Hashimoto, Yonatan Oren, and Percy Liang. 2018.
\newblock Generating sentences by editing prototypes.
\newblock \emph{Transactions of the Association of Computational Linguistics},
  6:437--450.

\bibitem[{Han and Baldwin(2011)}]{han2011lexical}
Bo~Han and Timothy Baldwin. 2011.
\newblock Lexical normalisation of short text messages: Makn sens a\# twitter.
\newblock In \emph{Proceedings of the 49th Annual Meeting of the Association
  for Computational Linguistics: Human Language Technologies-Volume 1}, pages
  368--378. Association for Computational Linguistics.

\bibitem[{Han et~al.(2012)Han, Cook, and Baldwin}]{han2012automatically}
Bo~Han, Paul Cook, and Timothy Baldwin. 2012.
\newblock Automatically constructing a normalisation dictionary for microblogs.
\newblock In \emph{Proceedings of the 2012 Joint Conference on Empirical
  Methods in Natural Language Processing and Computational Natural Language
  Learning}, pages 421--432. Association for Computational Linguistics.

\bibitem[{Hill et~al.(2016)Hill, Cho, Korhonen, and Bengio}]{hill2016learning}
Felix Hill, KyungHyun Cho, Anna Korhonen, and Yoshua Bengio. 2016.
\newblock Learning to understand phrases by embedding the dictionary.
\newblock \emph{{Transactions of the Association of Computational
  Linguistics}}, 4:17--30.

\bibitem[{Hochreiter and Schmidhuber(1997)}]{hochreiter1997long}
Sepp Hochreiter and J{\"u}rgen Schmidhuber. 1997.
\newblock Long short-term memory.
\newblock \emph{Neural Computation}, 9(8):1735--1780.

\bibitem[{Kingma and Ba(2015)}]{kingma2014adam}
Diederik~P. Kingma and Jimmy Ba. 2015.
\newblock Adam: {A} method for stochastic optimization.
\newblock In \emph{Proceedings of the 3rd International Conference on Learning
  Representations}.

\bibitem[{Kingma and Welling(2014)}]{kingma2013auto}
Diederik~P. Kingma and Max Welling. 2014.
\newblock Auto-encoding variational bayes.
\newblock In \emph{Proceedings of the 2nd International Conference on Learning
  Representations}.

\bibitem[{Knight et~al.(2006)Knight, Nair, Rathod, and
  Yamada}]{knight2006unsupervised}
Kevin Knight, Anish Nair, Nishit Rathod, and Kenji Yamada. 2006.
\newblock Unsupervised analysis for decipherment problems.
\newblock In \emph{Proceedings of the 21st International Conference on
  Computational Linguistics and 44th Annual Meeting of the Association for
  Computational Linguistics}, pages 499--506.

\bibitem[{Larsen et~al.(2016)Larsen, S{\o}nderby, Larochelle, and
  Winther}]{larsen2016autoencoding}
Anders Boesen~Lindbo Larsen, S{\o}ren~Kaae S{\o}nderby, Hugo Larochelle, and
  Ole Winther. 2016.
\newblock Autoencoding beyond pixels using a learned similarity metric.
\newblock In \emph{Proceedings of the 33nd International Conference on Machine
  Learning}, pages 1558--1566.

\bibitem[{Lin(2004)}]{lin2004rouge}
Chin-Yew Lin. 2004.
\newblock Rouge: A package for automatic evaluation of summaries.
\newblock \emph{Text Summarization Branches Out}.

\bibitem[{Liu et~al.(2012)Liu, Weng, and Jiang}]{liu2012broad}
Fei Liu, Fuliang Weng, and Xiao Jiang. 2012.
\newblock A broad-coverage normalization system for social media language.
\newblock In \emph{Proceedings of the 50th Annual Meeting of the Association
  for Computational Linguistics: Long Papers-Volume 1}, pages 1035--1044.
  Association for Computational Linguistics.

\bibitem[{Liu et~al.(2011)Liu, Weng, Wang, and Liu}]{liu2011insertion}
Fei Liu, Fuliang Weng, Bingqing Wang, and Yang Liu. 2011.
\newblock Insertion, deletion, or substitution?: normalizing text messages
  without pre-categorization nor supervision.
\newblock In \emph{Proceedings of the 49th Annual Meeting of the Association
  for Computational Linguistics: Human Language Technologies: short
  papers-Volume 2}, pages 71--76. Association for Computational Linguistics.

\bibitem[{Ni and Wang(2017)}]{ni2017learning}
Ke~Ni and William~Yang Wang. 2017.
\newblock Learning to explain non-standard english words and phrases.
\newblock In \emph{Proceedings of the Eighth International Joint Conference on
  Natural Language Processing (Volume 2: Short Papers)}, volume~2, pages
  413--417.

\bibitem[{Noraset et~al.(2017)Noraset, Liang, Birnbaum, and
  Downey}]{noraset2017definition}
Thanapon Noraset, Chen Liang, Larry Birnbaum, and Doug Downey. 2017.
\newblock Definition modeling: Learning to define word embeddings in natural
  language.
\newblock In \emph{Thirty-First AAAI Conference on Artificial Intelligence}.

\bibitem[{Papineni et~al.(2002)Papineni, Roukos, Ward, and
  Zhu}]{papineni2002bleu}
Kishore Papineni, Salim Roukos, Todd Ward, and Wei-Jing Zhu. 2002.
\newblock {BLEU}: a method for automatic evaluation of machine translation.
\newblock In \emph{Proceedings of the 40th Annual Meeting on Association for
  Computational Linguistics}, pages 311--318. Association for Computational
  Linguistics.

\bibitem[{Pavlopoulos et~al.(2017)Pavlopoulos, Malakasiotis, and
  Androutsopoulos}]{pavlopoulos2017deeper}
John Pavlopoulos, Prodromos Malakasiotis, and Ion Androutsopoulos. 2017.
\newblock Deeper attention to abusive user content moderation.
\newblock In \emph{Proceedings of the 2017 Conference on Empirical Methods in
  Natural Language Processing}, pages 1125--1135.

\bibitem[{Ravi and Knight(2011)}]{ravi2011deciphering}
Sujith Ravi and Kevin Knight. 2011.
\newblock Deciphering foreign language.
\newblock In \emph{Proceedings of the 49th Annual Meeting of the Association
  for Computational Linguistics: Human Language Technologies-Volume 1}, pages
  12--21. Association for Computational Linguistics.

\bibitem[{Rezende et~al.(2014)Rezende, Mohamed, and
  Wierstra}]{rezende2014stochastic}
Danilo~Jimenez Rezende, Shakir Mohamed, and Daan Wierstra. 2014.
\newblock Stochastic backpropagation and approximate inference in deep
  generative models.
\newblock In \emph{Proceedings of the 31st International Conference on
  International Conference on Machine Learning}, pages 1278--1286.

\bibitem[{Salimans et~al.(2015)Salimans, Kingma, and
  Welling}]{salimans2015markov}
Tim Salimans, Diederik Kingma, and Max Welling. 2015.
\newblock Markov chain monte carlo and variational inference: Bridging the gap.
\newblock In \emph{Proceedings of the 32nd International Conference on Machine
  Learning}, pages 1218--1226.

\bibitem[{Sohn et~al.(2015)Sohn, Lee, and Yan}]{sohn2015learning}
Kihyuk Sohn, Honglak Lee, and Xinchen Yan. 2015.
\newblock Learning structured output representation using deep conditional
  generative models.
\newblock In \emph{Advances in Neural Information Processing Systems 28: Annual
  Conference on Neural Information Processing Systems 2015}, pages 3483--3491.

\bibitem[{Sutskever et~al.(2014)Sutskever, Vinyals, and
  Le}]{sutskever2014sequence}
Ilya Sutskever, Oriol Vinyals, and Quoc~V Le. 2014.
\newblock Sequence to sequence learning with neural networks.
\newblock In \emph{Advances in Neural Information Processing Systems 27: Annual
  Conference on Neural Information Processing Systems 2014}, pages 3104--3112.

\bibitem[{Waseem and Hovy(2016)}]{waseem2016hateful}
Zeerak Waseem and Dirk Hovy. 2016.
\newblock Hateful symbols or hateful people? predictive features for hate
  speech detection on {Twitter}.
\newblock In \emph{Proceedings of the Student Research Workshop, SRW@HLT-NAACL
  2016}, pages 88--93.

\bibitem[{Wu et~al.(2016)Wu, Schuster, Chen, Le, Norouzi, Macherey, Krikun,
  Cao, Gao, Macherey et~al.}]{wu2016google}
Yonghui Wu, Mike Schuster, Zhifeng Chen, Quoc~V Le, Mohammad Norouzi, Wolfgang
  Macherey, Maxim Krikun, Yuan Cao, Qin Gao, Klaus Macherey, et~al. 2016.
\newblock Google's neural machine translation system: Bridging the gap between
  human and machine translation.
\newblock \emph{arXiv preprint arXiv:1609.08144}.

\bibitem[{Wulczyn et~al.(2017)Wulczyn, Thain, and Dixon}]{wulczyn2017ex}
Ellery Wulczyn, Nithum Thain, and Lucas Dixon. 2017.
\newblock Ex machina: {Personal} attacks seen at scale.
\newblock In \emph{Proceedings of the 26th International Conference on World
  Wide Web}, pages 1391--1399.

\bibitem[{Zhang et~al.(2016)Zhang, Xiong, Duan, Zhang
  et~al.}]{zhang2016variational}
Biao Zhang, Deyi Xiong, Hong Duan, Min Zhang, et~al. 2016.
\newblock Variational neural machine translation.
\newblock In \emph{Proceedings of the 2016 Conference on Empirical Methods in
  Natural Language Processing}, pages 521--530.

\bibitem[{Zhao et~al.(2017)Zhao, Zhao, and Eskenazi}]{zhao2017learning}
Tiancheng Zhao, Ran Zhao, and Maxine Eskenazi. 2017.
\newblock Learning discourse-level diversity for neural dialog models using
  conditional variational autoencoders.
\newblock In \emph{Proceedings of the 55th Annual Meeting of the Association
  for Computational Linguistics (Volume 1: Long Papers)}, volume~1, pages
  654--664.

\end{thebibliography}
\bibliographystyle{acl_natbib}
\appendix

\end{document}